# MAP Estimation, Linear Programming and Belief Propagation with Convex Free Energies

**Yair Weiss, Chen Yanover and Talya Meltzer**
School of Computer Science and Engineering
The Hebrew University of Jerusalem, Jerusalem, Israel
{*yweiss,cheny,talyam*}@cs.huji.ac.il

## Abstract

Finding the most probable assignment (MAP) in a general graphical model is known to be NP hard but good approximations have been attained with max-product belief propagation (BP) and its variants. In particular, it is known that using BP on a single-cycle graph or tree reweighted BP on an arbitrary graph will give the MAP solution if the beliefs have no ties.

In this paper we extend the setting under which BP can be used to provably extract the MAP. We define *Convex BP* as BP algorithms based on a convex free energy approximation and show that this class includes ordinary BP with single-cycle, tree reweighted BP and many other BP variants. We show that when there are no ties, fixed-points of convex max-product BP will provably give the MAP solution. We also show that convex sum-product BP at sufficiently small temperatures can be used to solve linear programs that arise from relaxing the MAP problem. Finally, we derive a novel condition that allows us to derive the MAP solution even if some of the convex BP beliefs have ties. In experiments, we show that our theorems allow us to find the MAP in many real-world instances of graphical models where exact inference using junction-tree is impossible.

## 1 Introduction

The task of finding the maximum aposteriori assignment (or MAP) in a graphical model comes up in a wide range of applications including image understanding [19], error correcting codes [3] and protein folding [27]. For an arbitrary graph, this problem is known to be NP hard [16] and various approximation algorithms have been proposed (see. e.g [13, 18] for recent reviews).

*Linear Programming (LP) Relaxations* are a standard method for approximating combinatorial optimization problems in computer science [1]. They have been used for approximating the MAP problem in a general graphical model by Santos [15]. More recently, LP relaxations have been used for error-correcting codes [3] and for protein folding [9]. LP relaxations have an advantage over other approximate inference schemes in that they come with an optimality guarantee – if the solution to the linear program is integer, then it is guaranteed to give the global optimum of the MAP problem.

The research described here is based on a remarkable recent set of results by Wainwright, Jaakkola and Willsky [21, 22] who discussed a variant of belief propagation called "tree reweighted belief propagation (TRBP)". They showed that when the TRBP output satisfied certain easy-to-check conditions, one could *provably* extract the MAP assignment from the TRBP output. Furthermore, they showed an intriguing connection between TRBP and LP relaxation.

In related work, we have used TRBP on a number of real world applications [26] and our experience with it raised a number of questions. First, TRBP is based on a distribution over spanning trees of the original graph. We wanted to know whether the properties of TRBP also hold for other BP variants that are not based on spanning trees. Second, in some applications the sufficient conditions given by Wainwright et al. [22] for extracting the MAP do not hold. We wanted to know whether one could extend these conditions.

In this paper, we show that the answer to both questions is affirmative. We define a family of algorithms called *convex BP* which refer to belief propagation with a convex free energy approximation. We show that tree reweighted BP suggested by Wainwright and colleagues [21] is a special case of convex BP but there are many convex free energies that cannot be represented as a tree reweighted free energy. This result has theoretical implications since it shows that the property of solving the LP is distinct from the property of providing a rigorous bound on the free energy, as well as practical implications since it provides an expanded family of possible LP algorithms.

We also discuss the max-product version of convex BP and show that when convex BP has beliefs without ties, the max-product assignment is guaranteed to be the MAP assignment. This gives a unified proof for previous results on ordinary BP with a single cycle [4, 20, 23] and tree reweighted BP [21]. Finally, we give a new theoretical condition that allows us to provably extract the MAP from convex BP beliefs, even if they have ties. We illustrate the power of these theorems on graphical models with hundreds of variables arising from computational biology and error-correcting codes.

## 1.1 MAP and LP relaxation

Given an observation vector $y$, we wish to perform inference on $\Pr(x|y)$ which is assumed to factorize into a product of potential functions:

$$\Pr(x|y) = \frac{1}{Z} \prod_\alpha \psi_\alpha(x_\alpha) = \frac{1}{Z} e^{-\sum \alpha E_\alpha(x_\alpha)}$$

where $\alpha$ is the domain of the potential $\psi_\alpha$ (the set of all variables that participate in the potential) and we define the "energy" $E_\alpha(x_\alpha)$ as the negative logarithm of the potential.

The MAP is the vector $x^*$ which maximizes the posterior probability:

$$x^* = \arg\max_x \prod_\alpha \psi_\alpha(x_\alpha) = \arg\min_x \sum_\alpha E_\alpha(x_\alpha)$$

To define the LP relaxation, we first reformulate the MAP problem as one of integer programming. We introduce indicator variables $q_i(x_i)$ for each individual variable and additional indicator variables $q_\alpha(x_\alpha)$ for all the potential domains. Using these indicator variables we define the integer program:

Minimize:
$$\sum_\alpha \sum_{x_\alpha} q_\alpha(x_\alpha) E_\alpha(x_\alpha)$$

Subject to:
$$\forall \alpha, x_\alpha \quad q_\alpha(x_\alpha) \in \{0, 1\}$$
$$\forall \alpha \quad \sum_{x_\alpha} q_\alpha(x_\alpha) = 1$$
$$\forall i, x_i, \alpha : i \in \alpha \quad \sum_{x_{\alpha \setminus i}} q_\alpha(x_\alpha) = q_i(x_i)$$

where the last equation, enforces the consistency of indicator variables for different potential domains.

This integer program is completely equivalent to the original MAP problem, and is hence computationally intractable. We can obtain the linear programming relaxation by allowing the indicator variables to take on non-integer values. That is, we replace the constraint $q_\alpha(x_\alpha) \in \{0, 1\}$. with $q_\alpha(x_\alpha) \in [0, 1]$. This problem can now be solved efficiently, and if the solutions to the LP happen to be integer, we have provably found the MAP.

## 1.2 Belief Propagation and its variants

As shown by Yedidia et al. [28], there exist a large number of free energy approximations that are based on a set of "double counting numbers". These double counting numbers are used to approximate the entropy of $x$, denoted $\tilde{H}$, by means of a linear combination of entropies over individual variables $i$, $H_i$, and variables that participate in a factor $\alpha$, $H_\alpha$:

$$\tilde{H} = \sum_\alpha c_\alpha H_\alpha + \sum_i c_i H_i$$

Given a set of double counting numbers $c_\alpha, c_i$ we define the approximate free energy functional. This is a functional that takes as input a set of approximate marginals $b_\alpha(x_\alpha), b_i(x_i)$ and uses them to define the average energy and the approximate entropy. The approximate free energy at temperature $T$ is simply:

$$F(b_\alpha, b_i) = U(b_\alpha) - T\tilde{H}(b_\alpha, b_i) \qquad (1)$$

where the average energy, $U(b_\alpha)$, and the approximate entropy, $\tilde{H}(b_\alpha, b_i)$, are given by:

$$U(b_\alpha) = \sum_\alpha \sum_{x_\alpha} b_\alpha(x_\alpha) E_\alpha(x_\alpha)$$
$$\tilde{H}(b_\alpha, b_i) = \sum_\alpha c_\alpha \sum_{x_\alpha} b_\alpha(x_\alpha) \ln b_\alpha(x_\alpha)$$
$$+ \sum_i c_i \sum_{x_i} b_i(x_i) \ln b_i(x_i)$$

A special case of approximate free energies is when $c_i = 1 - d_i, c_\alpha = 1$, where $d_i$ is the number of factors that node $i$ participates in (or equivalently, the degree of node $i$ in the factor graph). In this case the approximate free energy is called the Bethe free energy.

Given an approximate free energy, there are many possible algorithms that try to minimize it. For concreteness we give here one possible algorithm, the two-way GBP algorithm [28], but we should emphasize that all our results hold for any algorithm that converges to stationary points of the approximate free energy (e.g. [21, 24, 25]). Assuming $c_\alpha = 1$ for all factors, the two-way algorithm is similar to ordinary BP on a factor graph, but with an additional "reweighting" step. As in ordinary BP, we denote the messages sent from factor node $\alpha$ to variable node $i$ by $m_{\alpha i}(x_i)$ and the message from variable node $i$ to factor node $\alpha$ by $m_{i\alpha}(x_i)$. The messages are updated as follows:

$$m_{\alpha i}^0(x_i) = \sum_{x_{\alpha\setminus i}} \psi_\alpha^{1/T}(x_\alpha) \prod_{j\neq i} m_{j\alpha}(x_j)$$

$$m_{i\alpha}^0(x_i) = \prod_{\beta\neq\alpha} m_{\beta i}(x_i)$$

$$m_{\alpha i}(x_i) \leftarrow \left(m_{\alpha i}^0(x_i)\right)^{\gamma_i} \left(m_{i\alpha}^0(x_i)\right)^{\gamma_i - 1}$$

$$m_{i\alpha}(x_i) \leftarrow \left(m_{i\alpha}^0(x_i)\right)^{\gamma_i} \left(m_{\alpha i}^0(x_i)\right)^{\gamma_i - 1}$$

with $\gamma_i = \frac{deg(i)}{1-c_i}$. The max-product belief-propagation algorithm is the same, but with the sum replaced with a max. Note that when $\gamma_i = 1$ (or, equivalently, $c_i = 1 - deg(i)$) the above update equations reduce to ordinary-BP. From the messages we calculate the beliefs:

$$b_i(x_i) \propto \prod_\alpha m_{\alpha i}(x_i)$$

$$b_\alpha(x_\alpha) \propto \psi_\alpha^{1/T}(x_\alpha) \prod_i m_{i\alpha}(x_i) \quad (2)$$

We emphasize again that this is just one possible algorithm to find stationary points of the approximate free energy. In order to deal with any algorithm, we use the following characterization of approximate free energy stationary points. This characterization follows directly from differentiating the Lagrangian of the approximate free energy and was used by [8, 21, 29].

*Observation:* A set of beliefs $b_\alpha, b_i$ are stationary points of an approximate free energy with double counting numbers $c_\alpha, c_i$ and temperature $T$ if and only if they satisfy:

- **Admissibility:** for all $x$:

$$(\Pr(x))^{1/T} \propto \prod_\alpha b_\alpha^{c_\alpha}(x_\alpha) \prod_i b_i^{c_i}(x_i) \quad (3)$$

- **Marginalization:** The beliefs are positive, sum to one and satisfy:

$$\forall i, \alpha : i \in \alpha \quad \sum_{x_{\alpha\setminus i}} b_\alpha(x_\alpha) = b_i(x_i) \quad (4)$$

Similarly, it can be shown that a set of beliefs are fixed-points of the max-product algorithm with double counting numbers $c_\alpha, c_i$ if and only if they satisfy the above admissibility condition and **max-marginalization** condition:

$$\forall i, \alpha : i \in \alpha \quad \max_{x_{\alpha\setminus i}} b_\alpha(x_\alpha) = b_i(x_i) \quad (5)$$

In summary, we have defined the MAP problem, the LP relaxation and a family of belief propagation algorithms. The natural questions that arise are:

- When can BP algorithms be used to solve the LP relaxation?
- How are the max-product and sum-product algorithms related?
- When can BP algorithms be used to provably extract the MAP assignment?

## 2 Convex Free energies

As we will show subsequently, a key property in analyzing approximate free energies is their convexity over the set of constraints.[1] Heskes [6, 7] has derived sufficient conditions for an entropy approximation to be convex over the set of constraints. In our setting, we can rewrite these conditions as follows:

**Definition:** An approximate entropy term of the form:

$$H = \sum_\alpha c_\alpha H_\alpha + \sum_i c_i H_i \quad (6)$$

is said to be *provably convex* if there exist *non-negative* numbers $c_{i\alpha}, d_\alpha, d_i$ such that:

$$H = \sum_{i,\alpha: i\in\alpha} c_{i\alpha}(H_\alpha - H_i) + \sum_\alpha d_\alpha H_\alpha + \sum_i d_i H_i$$

### 2.1 Tree Reweighted Free Energies

Wainwright and colleagues have introduced an important subclass of belief propagation algorithms: *tree reweighted BP*. These are algorithms whose free energy is a linear combination of free energies defined on spanning trees of the graph. They have shown that tree reweighted BP (1) can be used to obtain a rigorous bound on the free energy and (2) gives rise to a convex free energy approximation. A natural question that arises is whether these two properties of TRBP are equivalent – do all BP algorithms that arise from convex free energies also give a rigorous bound on the free energy. In this section we show that the answer is negative. In fact tree reweighted BP algorithms represent a small fraction of convex free energy belief propagation algorithms.

---

[1]Convexity over the set of constraints means the function is convex as a function of any beliefs that satisfy the marginalization constraints. This is a weaker assumption from convexity over any beliefs. Henceforth we refer to this weaker assumption as convexity of the entropy approximation.

Tree-reweighted free energies [21] use entropy terms of the form:

$$H_{TRBP}(\mu) = \sum_{\mathcal{T}} \mu_{\mathcal{T}} H_{\mathcal{T}} \qquad (7)$$

where $\mathcal{T}$ is a spanning tree in the graph, $\mu_{\mathcal{T}}$ defines a distribution over spanning trees and $H_{\mathcal{T}}$ is the entropy of that tree. Since $H_{\mathcal{T}}$ is convex, so is $H_{TRBP}$. But not every convex free energy can be written in this way. To see this, note that any tree reweighted entropy can be rewritten:

$$H_{TRBP}(\mu) = \sum_{<ij>} \rho_{ij} H_{ij} + \sum_i (1 - \sum_j \rho_{ij}) H_i$$

where $\rho_{ij}$ is the edge appearance probability defined by $\mu$. In comparing this to the general entropy approximation (equation 6) we see that tree reweighted entropies are missing a degree of freedom (with $c_i$). In fact, for any TRBP entropy we can add an infinite number of possibile positive combination of single node entropies and still maintain convexity. Thus, TRBP entropies are a measure zero set of all convex entropies.

In some cases, we can even subtract single node entropies from a TRBP entropy and still maintain convexity. For example, the Bethe free energy for a single cycle can be shown to be convex but *it cannot be represented as tree-reweighted free energy* [21]. In particular, it does not give rise to a bound on the free energy.

This shows that the family of BP algorithms that provide a bound on the free energy is a strict subset of the family of convex BP algorithms.

## 3 When does sum-product BP solve the LP relaxation?

**Claim: Convex BP=LP** Let $b_\alpha, b_i$ be fixed-point beliefs from running belief propagation with a convex entropy approximation at temperature $T$. As $T \to 0$ these beliefs approach the solution to the linear program.

**Proof:** We know that the BP beliefs are constrained stationary points of the free energy (equation 1). The minimization of $F$ is done subject to the following constraints:

$$\begin{aligned} b_\alpha(x_\alpha) &\in [0,1] \\ \sum_{x_\alpha} b_\alpha(x_\alpha) &= 1 \\ \sum_{x_{\alpha \setminus i}} b_\alpha(x_\alpha) &= b_i(x_i) \end{aligned}$$

The energy term is exactly the LP problem. As we decrease the temperature, the approximate free energy approaches the LP cost (note that the entropy term is bounded). If we assume the entropy function to be convex then the approximate free energy is convex and hence any fixed-point corresponds to the global minimum. □

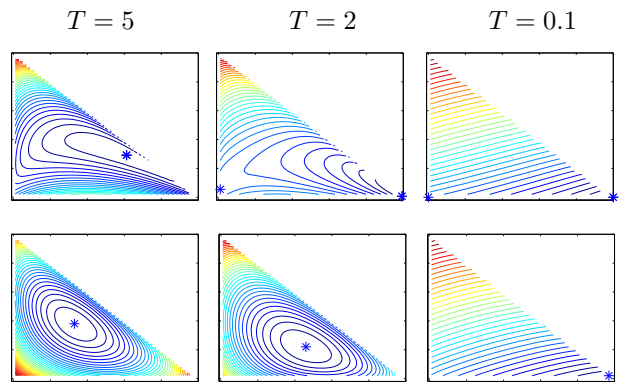

Figure 1: Contour plots of the Bethe free energy (top) and a convex free energy (bottom) for a 2D Ising model with uniform external field at different temperatures. The stars indicate local stationary points. Both free energies approach the LP as temperature is decreased, but for the Bethe free energy, a local minimum is present even for arbitrarily small temperatures.

Note that for any BP algorithm, it is true that the approximate free energy minimization problem approaches the LP problem. In particular, this is true for ordinary BP which minimizes the Bethe free energy. However, when the entropy function is non-convex, there is no guarantee that fixed-points will correspond to the global optimum.

Figure 1 illustrates the difference. We consider a graphical model corresponding to a toroidal grid. The nodes are binary and all the pairwise potentials are of the form:

$$\Psi = \begin{pmatrix} 3 & 1 \\ 1 & 2 \end{pmatrix}$$

These potentials correspond to an Ising model with a uniform external field – nodes prefer to be similar to their neighbors and there is a preference for one state over the other. In order to visualize the approximate free energies, we consider beliefs that are symmetric and identical for all pairs of nodes:

$$b_\alpha = \begin{pmatrix} x & y \\ y & 1-(x+2y) \end{pmatrix}$$

Note that the MAP (and the optimum of the LP) occur at $x=1, y=0$ in which case all nodes are in their preferred state. Figure 1 shows the Bethe free energy (top) and a convex free energy (bottom) for this problem for different temperatures. The stars indicate local stationary points. Both free energies approach the LP as temperature is decreased, but for the Bethe free energy, a local minimum is present even for arbitrarily small temperatures.

## 4 How are max-product BP and sum-product BP related?

Although we have shown that one can use sum-product convex BP to solve the linear program, one needs to be able to run the sum-product algorithm at sufficiently small temperatures and this may cause serious numerical problems. We now show that in certain cases, one can solve the linear program by running the max-product algorithm at any temperature. This follows from the interpertation of the max-product algorithm as the zero temperature limit of sum-product.

**Zero temperature lemma:** Suppose $\{b_\alpha(x_\alpha), b_i(x_i)\}$ are fixed-points of the sum-product algorithm at temperature $T$. Define $\hat{b}_\alpha(x_\alpha) \propto b_\alpha^T(x_\alpha)$ and $\hat{b}_i(x_i) \propto b_i^T(x_i)$. Then for any $T \to 0$, $\{\hat{b}_\alpha(x_\alpha), \hat{b}_i(x_i)\}$ approach the conditions for fixed-points of the max-product BP algorithm at temperature $T = 1$.

**Proof:** Recall that a set of beliefs are fixed-points of the sum-product algorithm if and only if they satisfy the admissibility constraint (equation 3) and the marginalization constraint (equation 4) and they are fixed-points of the max-product algorithm if and only if they satisfy the admissibility constraint and the max-marginalization constraint (equation 5).

For any $T$, if $\{b_\alpha(x_\alpha), b_i(x_i)\}$ satisfy the admissibility constraint at temperature $T$ then $\hat{b}_\alpha(x_\alpha) \propto b_\alpha^T(x_\alpha)$ and $\hat{b}_i(x_i) \propto b_i^T(x_i)$ must satisfy the admissibility constraint at temperature 1. We just need to show that $\hat{b}_\alpha, \hat{b}_i$ also satisfy the max-marginalization constraint as $T \to 0$. Since $\{b_\alpha(x_\alpha), b_i(x_i)\}$ are fixed-points of the sum-product algorithm, they must satisfy sum-marginalization, and substituting in the definition of $\{\hat{b}_\alpha(x_\alpha), \hat{b}_i(x_i)\}$ we obtain:

$$\sum_{x_{\alpha\setminus i}} \hat{b}_\alpha^{1/T}(x_\alpha) = \hat{b}_i^{1/T}(x_i)$$

This can be rewritten:

$$\left(\sum_{x_{\alpha\setminus i}} \hat{b}_\alpha^{1/T}(x_\alpha)\right)^T = \hat{b}_i(x_i)$$

and as $T \to 0$ this approaches the max-marginalization constraint. □

The zero temperature lemma suggests that if we could run sum-product BP at arbitrarily small temperatures, we could use the beliefs to find fixed-points of max-product BP at temperature $T = 1$. But to use max-product BP to solve the LP we want to go in the opposite direction, i.e. use max-product BP to define fixed-points of sum-product BP at small temperatures. It turns out that this direction does not always work, as the following counterexample shows.

**Example**: Consider a graphical model with two nodes $x_1, x_2$ and a pairwise factor:

$$\psi_{12}(x_1, x_2) = \begin{pmatrix} 1 & 1 \\ 1 & 0 \end{pmatrix}.$$

Consider the Bethe approximation for this graph ($c_{12} = 1, c_1 = c_2 = 0$). This entropy approximation is trivially convex. It is easy to show that *for any temperature $T$*, the fixed-points of sum-product BP are:

$$b_{12} = \tfrac{1}{3}\begin{pmatrix} 1 & 1 \\ 1 & 0 \end{pmatrix}, \quad b_1 = b_2 = \tfrac{1}{3}(2,1).$$

And, again, *for any temperature $T$*, the fixed-points of max-product BP are:

$$b_{12} = \begin{pmatrix} 1 & 1 \\ 1 & 0 \end{pmatrix}, \quad b_1 = b_2 = (1,1).$$

In other words, when we run max-product BP we will get *uniform* beliefs in both nodes and no matter how small we set $T$, raising the beliefs to the power $1/T$ will still give uniform beliefs. However, the sum-product beliefs are non-uniform for any temperature. Note, however, that the counterexample still satisfies the zero temperature lemma — raising the sum-product beliefs to the power $T$ indeed approaches the max-product beliefs as $T \to 0$.

The counterexample shows the problem with going from max-product beliefs to sum-product beliefs at $T \to 0$ (which are equivalent to the LP solution) – the max-product beliefs retain the information on the maximum belief, but have lost the information regarding the *number of configurations* that attained the maximal value. This motivates the following sufficient conditions for going from max-product beliefs to sum-product beliefs.

Given a set of beliefs, $\hat{b}_\alpha, \hat{b}_i$ we define the *sharpened beliefs* as follows:

$$q_\alpha(x_\alpha) \propto \delta(\hat{b}_\alpha(x_\alpha) - \max_{x_\alpha} \hat{b}_\alpha(x_\alpha))$$
$$q_i(x_i) \propto \delta(\hat{b}_i(x_i) - \max_{x_i} \hat{b}_i(x_i))$$

To illustrate this definition, a belief vector $(0.6, 0.4)$ would be sharpened to $(1, 0)$ and a belief vector $(0.4, 0.4, 0.2)$ would be sharpened to $(0.5, 0.5, 0)$.

Using this definition it can be shown:

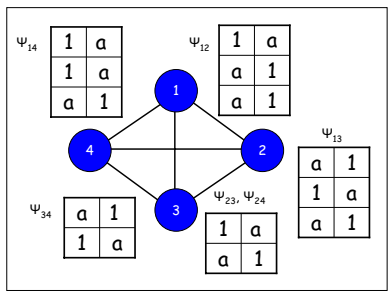

Figure 2: A simple problem for which max-product convex BP will converge in a single iteration but the beliefs cannot be used to solve the linear program. $a$ is a real number smaller than 1.

**Corollary: Max-Product Convex BP=LP** Let $\hat{b}_\alpha, \hat{b}_i$ be max-product beliefs at $T = 1$ for a convex BP algorithm. If the sharpened max-product beliefs are sum-marginalizable then they are a solution to the LP problem.

In the simple two node example, the sharpened max-product beliefs are simply the original beliefs, so they are not a solution to the LP problem. In this case, however, it is easy to "fix" the beliefs by defining $b_1, b_2$ as the sum marginals of $b_{12}$. Figure 2 shows a simple problem for which the problem is much harder to fix. Max-product convex BP will converge in a single iteration to beliefs that are proportional to the potentials, but the sharpened beliefs will not be sum marginalizeable. Hence they cannot be used to solve the linear program. However, sum-product convex BP at $T = 0.0001$ gave a solution to the LP.

To summarize, our analysis (as well as that by Kolmogorov and Wainwright [10, 11]) shows that the relation between LP relaxation and max-product convex BP is subtle – although we can always verify post-hoc whether we have obtained the LP solution, and a fixed-point corresponding to the LP solution is guaranteed to exist, we are not guaranteed to find that fixed-point. On the other hand, for sum-product convex BP the connection to LP is much more direct – at sufficiently small temperatures the BP beliefs will approach the LP solution.

## 5 When can we extract the MAP from max-product convex BP?

Whereas the previous section focused on using the max-product algorithm to avoid the numerical instabilities associated with sum-product at small temperatures, here we show how to use max-product BP directly to obtain a solution to the MAP problem.

**Theorem 1: Convex-BP = MAP without frustrations:** Let $b_\alpha, b_i$ be fixed-points of max-product BP with a provably convex entropy function. If there exists an assignment $x^*$ such that $b_\alpha(x^*_\alpha)$ maximizes $b_\alpha(x_\alpha)$ and $b_i(x^*_i)$ maximizes $b_i(x_i)$ then $x^*$ is the MAP.

**Proof:** Since we have fixed-points of max-product BP they are admissible (equation 3). Using the fact that the entropy is provably convex, we can rewrite this as:

$$\Pr(x) \propto \prod_{i,\alpha} \left(\frac{b_\alpha(x_\alpha)}{b_i(x_i)}\right)^{c_{i\alpha}} \prod_\alpha b_\alpha^{d_\alpha}(x_\alpha) \prod_i b_i^{d_i}(x_i) \quad (8)$$

We have rewritten $\Pr(x)$ as a product of functions on $x_\alpha, x_i$. We want to show that $x^*_\alpha, x^*_i$ maximize all of these functions. We know that $x^*_\alpha$ maximizes $b_\alpha(x_\alpha)$ and $x^*_i$ maximizes $b_i(x_i)$. Therefore, we just need to worry about the quotients:

$$r_{i\alpha}(x_\alpha) \equiv \frac{b_\alpha(x_\alpha)}{b_i(x_i)}$$

**Lemma:** Suppose we have a set of beliefs $b_\alpha, b_i$ that are max-marginalizable and there exists $x^*$ such that $b_\alpha(x^*_\alpha)$ maximizes $b_\alpha(x_\alpha)$ and $b_i(x^*_i)$ maximizes $b_i(x_i)$. Then $x^*$ also maximizes $b_\alpha(x_\alpha)/b_i(x_i)$.

This lemma was proved for the case of pairwise factors in [20] and the generalization for arbitrary factors is straightforward.

Using the Lemma we see that $x^*$ maximizes all the terms in the decomposition (equation 8), since each term is either $b_i(x_i)$, $b_\alpha(x_\alpha)$ or $r_{i\alpha}(x_i, x_{\alpha \setminus i})$, raised by the power of a non-negative number. $\square$

**Corollary Convex BP = MAP without ties** Let $b_\alpha, b_i$ be fixed-points of max-product BP with a provably convex entropy function. If there are no ties in these beliefs – for every $i$ the maximum of $b_i(x_i)$ is attained at a unique value $x^*_i$ – then $x^*$ is the MAP.

**Proof:** Since the beliefs are max-marginalizable the fact that there are no ties in the node beliefs implies there are no ties in the factor beliefs. It follows that $x^*_\alpha$ maximizes $b_\alpha(x_\alpha)$ for each $\alpha$ and hence the previous theorem holds.

Both the previous theorem and the corollary were proven for the case of TRBP by [21]. Our proof extends these results for *arbitrary* convex BP algorithms.

### 5.1 Dealing with frustrations

There are many cases in which it is impossible to find an assignment $x^*$ that maximizes all the factor beliefs. This happens whenever the beliefs define a frustrated cycle (see figure 3).

Our final theorem shows that it is possible to extract the MAP from convex BP beliefs even if there are frustrated cycles.

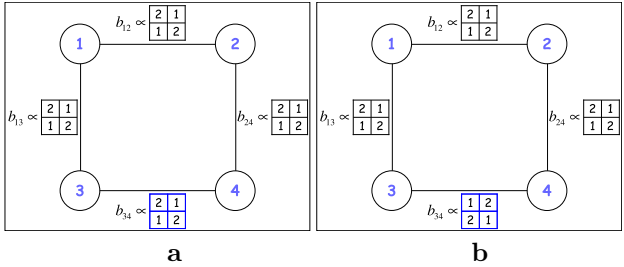

Figure 3: An illustration of a frustrated cycle. The tables show pairwise beliefs obtained by a convex BP algorithm. For the four nodes in (a), it is possible to find an assignment $x^*$ that maximizes the pairwise and singleton beliefs. Theorem 1 proves that this means that $x^*$ is the global optimum. For the four nodes in (b) it is impossible to find such an assignment.

**Theorem 2: Convex-BP = MAP with frustrations:** Let $b_\alpha, b_i$ be fixed-points of max-product BP with a provably convex entropy function. Let $NT$ be the set of non-tied variables and $T$ be the set of tied variables. We denote the set of tied nodes that have non-tied neighbors as $\partial T$. Define $x_{NT}^*$ by maximizing the local beliefs (the maximum here is unique since these are non-tied nodes). Define:

$$b_T(x_T) = \prod_{\substack{i,\alpha: \\ i \in T \setminus \partial T}} r_{i\alpha}^{c_{i\alpha}}(x_\alpha) \prod_{\alpha \subset T} b_\alpha^{d_\alpha}(x_\alpha) \prod_{i \in T \setminus \partial T} b_i^{d_i}(x_i)$$

If there exists $x_T^*$ that maximizes $b_T(x_T)$ and for all regions $\alpha$ that contain both tied and non-tied nodes $b_\alpha(x_\alpha^*)$ maximizes $b_\alpha(x_\alpha)$ then the assignment $(x_T^*, x_{NT}^*)$ is the MAP assignment.

**Proof:** Using the decomposition equation (equation 8) we can write:

$$\begin{aligned}
\Pr(x) &\propto \prod_{i,\alpha} r_{i\alpha}^{c_{i\alpha}}(x_\alpha) \prod_\alpha b_\alpha^{d_\alpha}(x_\alpha) \prod_i b_i^{d_i}(x_i) \\
&= b_T(x_T) \cdot \prod_{i,\alpha: i \notin T} r_{i\alpha}^{c_{i\alpha}}(x_\alpha) \prod_{i,\alpha: i \in \partial T} r_{i\alpha}^{c_{i\alpha}}(x_\alpha) \\
&\quad \cdot \prod_{\alpha \not\subset T} b_\alpha^{d_\alpha}(x_\alpha) \prod_{i \notin T} b_i^{d_i}(x_i) \prod_{i \in \partial T} b_i^{d_i}(x_i)
\end{aligned}$$

We want to show that $x^*$ maximizes all the terms in the decomposition. By construction $x_T^*$ maximizes $b_T(x_T)$. By the previous lemma, for all regions $\alpha \not\subset T$, $x_\alpha^*$ maximizes $r_{i\alpha}(x_\alpha)$. Also, for all $i \notin T$ and $\alpha \not\subset T$, $b_i(x_i)$ and $b_\alpha(x_\alpha)$ are maximized by $x_{NT}^*$. For $i \in \partial T$, $x_T^*$ maximizes $b_i(x_i)$ due to the assumption that $x^*$ maximizes the boundary factors. The only thing to worry about are terms of the sort $r_{i\alpha}(x_\alpha)$, where $i$ is a boundary node. But since the beliefs $b_\alpha, b_i$ are max-marginalizable for boundary nodes as well and $b_\alpha(x_\alpha^*)$ maximizes $b_\alpha(x_\alpha)$ by assumption, $b_i(x_i^*)$ must maximize $b_i(x_i)$ (otherwise this would contradict max-marginalizability). Hence we can use the lemma again to show that $x_\alpha^*$ must maximize $r_{i\alpha}(x_\alpha)$. □

**Corollary:** For pairwise factors, if all nodes on the boundary of the tied nodes, $\partial T$, have uniform beliefs, then the non-tied beliefs are optimal (that is, $x_{NT}^* = x_{NT}^{MAP}$).

This is because for uniform beliefs on the boundary, any assignment $x_T$ maximizes the beliefs on the boundary. The fact that factors are pairwise means that it also maximizes all factors that include the boundary nodes. This generalizes a result of Kolmogorov and Wainwright [11] for binary nodes.

## 6 An illustrative example

To illustrate the relationship between linear programming (LP), ordinary belief propagation (BP), tree reweighted belief propagation (TRBP) and convex belief propagation (CBP), we conducted simulations with a small grid graphical model – 9 nodes, arranged in a $3 \times 3$ grid.

|  BP   |   |   | TRBP |   |   | Default CBP |   |   | Trivial CBP |   |   |
|---|---|---|---|---|---|---|---|---|---|---|---|
| 1 | 2 | 1 | 0 | 1 | 0 | 1 | $\frac{3}{2}$ | 1 | 0 | 0 | 0 |
| 2 | 3 | 2 | 1 | 2 | 1 | $\frac{3}{2}$ | 2 | $\frac{3}{2}$ | 0 | 0 | 0 |
| 1 | 2 | 1 | 0 | 1 | 0 | 1 | $\frac{3}{2}$ | 1 | 0 | 0 | 0 |

Figure 4: Negative double counting numbers $-c_i$ for four different free energy approximations on a $3 \times 3$ grid used to illustrate the different algorithms.

One of the difficulties in comparing these different variants of belief propagation comes from the fact that there are many ways to construct TRBP or convex BP approximations. We define the default convex BP approximation based on the following observation.

**Observation:** For any factor graph, the free energy approximation given by $c_\alpha = 1$ and $c_i = -\sum_{\alpha:i\in\alpha} \frac{1}{d_\alpha}$ is convex.

This follows from the convexity decomposition in section 2 with $c_{i\alpha} = \frac{1}{d_\alpha}$, $d_i = 0$ and $d_\alpha = 0$, where $d_\alpha$ is the number of nodes that participate in the factor $\alpha$.

We consider 4 different approximate free energies which give the double counting numbers $c_i$ in figure 4. In all of them, all the factors have the same double counting number $c_\alpha = 1$ but they differ in the double counting numbers $c_i$ for the nodes. In ordinary BP, $c_i = 1 - d_i$. For TRBP we considered two spanning forests – one for the horizontal edges of the grid, and one for the vertical edges. We used the uniform distribution over these two spanning forests so that the edge appearance probability was 0.5 for all edges. To facilitate comparison with the other approximations, we multiplied the entropy approximation by two, so that $c_\alpha = 1$ for all edges and $c_i = 2 - deg(i)$ for all nodes. We also considered two convex BP approximations. The default CBP approximation gives $c_i = -d_i/2$ (since all the factors are pairwise). Finally the trivial approximation $c_i = 0$ is trivially convex since it only sums up positive entropies. For all these free energy approximations we ran the max-product algorithm with a-synchronous updates and a "dampening" factor of 0.5.

We generated 100 samples of these $3 \times 3$ "spin glasses" – the energy was given by $E(x) = \sum_i J_{ii} x_i + \sum_{<ij>} J_{ij} x_i x_j$. $J_{ii}$ and $J_{ij}$ were sampled from zero mean Gaussians with standard deviations 0.4 and 1.0.

We found that the problems could be subdivided into three classes based on the behavior of the linear programming relaxation. In the **easy regime** the linear programming solution is all integer and hence solving the LP gives the MAP (this happened in 53% of our runs). All the convex approximations converged to beliefs without ties. Consistent with the *Max-Product Convex BP=LP* corollary, the assignments obtained by all the approximations were indeed the MAP. Additionally, in all the simulations in the easy regime, ordinary BP gave the correct answer. However, whereas the convex algorithms come with a MAP certificate, ordinary BP comes with no such theoretical guarantee. While all algorithms found the right answer in this easy regime, the number of iterations to convergence was different. Ordinary BP converged faster (median number of iterations 48), then the default CBP (112 iterations), then TRBP (176 iterations) and finally the trivial CBP (225 iterations).

In the **hard regime** the LP solution is all fractional (this happened in 36% of the runs). Consistent with the *Max-Product Convex BP=LP* corollary, all the convex BP algorithms converged in this case to beliefs where all the nodes were tied. For this regime, ordinary BP *never converged*. Although the zero-temperature lemma guarantees that a fixed-point with ties exists for ordinary BP as well, this fixed-point was never found. In this regime, TRBP converged the fastest among the convex BP algorithms (median number of iterations 185), followed by the trivial CBP (295 iterations) and finally the default CBP (316 iterations). However, in terms of finding the MAP, all convex algorithms were equally useless.

In the **intermediate regime** the LP solution is partially integer and partially fractional (this happened in 11% of the runs). Again, all the convex BP algorithms converged to the same solution where part of the beliefs are tied and others are not (tied beliefs corresponded to fractional solutions to the LP). The default CBP was fastest (144 iterations) followed by TRBP (197 iterations) and then trivial CBP (372). In the majority of these cases (8 out of 11), ordinary BP did not converge.

To summarize, convex BP algorithms have the greatest practical advantage over ordinary BP in the intermediate regime where the LP is partially fractional - they converge better and allow to provably extract the MAP. All convex BP algorithms are equivalent in terms of finding the MAP but convergence rate can vary drastically.

## 7 Real World Experiments

The experiments reported here were designed to see how often convex BP will allow us to solve real-world instances of the MAP problem. Checking the conditions of theorems 1 and 2 requires finding the MAP in a reduced graphical model defined over the tied nodes. We use the junction tree algorithm to solve this task so this becomes infeasible when the subgraph of tied nodes has large induced width.

Our first two datasets are based on real-world graphical models coming from computational biology. We briefly summarize the construction of these datasets (see [26] for more details).

Proteins are chains of *residues*, each containing one of 20 possible amino acids. All amino acids are connected together by a common backbone structure, onto which amino-specific side-chains are attached. The problem of predicting the residue side-chain conformations given a backbone structure is considered of central importance in protein-folding and molecular design and has been tackled extensively using a wide variety of methods (for a recent review, see [2]). The typical way to predict side-chain configurations is to define an energy function and a discrete set of possible side-chain conformations, and then search for the minimal energy configuration. Even when the energy function contains only pairwise interactions, the configuration space grows exponentially and it can be shown that the prediction problem is NP-complete [5].

As a dataset we used 370 X-ray crystal structures with resolution better than or equal to 2Å, R factor below 20% and mutual sequence identity less than 50%. Each protein consist of a single chain and up to 1,000 residues. Protein structures were acquired from the Protein Data Bank site (http://www.rcsb.org/pdb). For each protein, we have built a graphical model using the ROSETTA energy function [12]. The nodes of this model correspond to residues, and there are edges between any two residues that interact [27]; the potentials are inversely related to the energy.

| Task | LP=IP | Thm 1 | Thm 2 | Failed |
|---|---|---|---|---|
| Side-Chain | 1.35% | 83.78% | 6.76% | 8.1% |
| Design | 0% | 2.1% | 0% | 97.9% |

Table 1: The percentage of real-world instances solved by the different theorems presented in this paper. LP=IP means that the solution of the LP was integer. For the easier problem of side-chain prediction (top) we could find the global optimum for about 92% of the cases. For the harder task of protein design, there are so many tied nodes that checking the conditions of the theorems becomes infeasible.

The *protein design* problem is the inverse of the protein folding problem. Given a particular 3D shape, we wish to find a sequence of amino-acids that will be as stable as possible in that 3D shape. Typically this is done by finding a set of (1) amino-acids and (2) rotamer configurations that minimize an approximate energy [17]. While the protein design problem is quite different from side-chain prediction it can be solved using the same graph structure. The only difference is that now the nodes do not just denote rotamers but also the identity of the amino-acid at that location. Thus, the state-space here is significantly larger than in the side-chain prediction problem. We, again, used the ROSETTA energy function to define the pairwise and local potentials. As a dataset we used 96 X-ray crystal structures, 40-180 amino acids long. For each of these proteins, we allowed all residues to assume any rotamer of any amino acid. There are, therefore, hundreds of possible states for each node.

We found that convergence was not an issue – in all the experiments convex BP converged in reasonable time but the number of ties determines the success of the algorithm. Table 1 shows a breakdown of the success rate for the two problems. In the harder problem of protein design, the number of ties is so large that checking the conditions of theorems 1 and 2 is infeasible. But in the side-chain problem, even though exact inference is NP hard and the search space can be as large as $10^{600}$ (largest clique in junction tree – $10^{60}$), the number of ties is quite manageable – in over 90% of the instances we can find the global optimum. For this data set, ordinary BP also converged 85.14% of the times, and whenever it converged it found the global optimum.

Our third dataset was based on a low density parity check code taken from David Mackay's encyclopedia of sparse graph codes (`http://www.inference.phy.cam.ac.uk/mackay/codes/data.html`). We used the 204.33.484 code which has 204 bits and 102 parity checks. We simulated sending a codeword over a binary symmetric channel. Each received word defines the local factors in a factor graph, and we used trivial convex BP to find the MAP in this graphical model. We repeated this experiment for different signal to noise ratios (SNR).

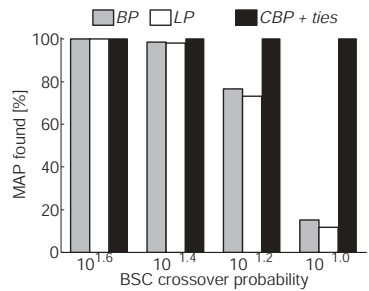

Figure 5: A comparison of success rates as a function of crossover probability for a LDPC code on a binary symmetric channel.

Figure 5 shows our results. For high SNR, the problem is easy and the LP solution is almost always integer (success of LP corresponds to a fully integer LP solution or, equivalently, max-product convex BP having no ties). However, as the SNR decreases, the LP solution is almost always partially fractional but using theorem 1 allows us to find the MAP decoding in all cases. Thus even though the search space here is of size $2^{204}$ and the maximal clique in the junction-tree includes 134 bits, convex BP allows us to find the global optimum in a matter of minutes.

## 8 Discussion

Belief Propgation and its variants have shown excellent performance as approximate inference algorithms. In this paper we have focused on conditions under which the MAP can be provably extracted from BP beliefs. We have shown that previous results – BP on a single cycle and TRBP on arbitrary graphs – are special cases of a wider result on BP with a convex free energy. We have also shown that BP with a convex free energy can be used to solve LP relaxations of the MAP problem. Finally, we have proven a novel result that allows us to extract the MAP from convex BP beliefs even when there are frustrated cycles.

From a theoretical perspective, one intriguing result arising from our work is the close connection between LP relaxations and a large class of belief propagation variants (including ordinary BP). Given the large amount of literature on the tightness of LP relaxations for combinatorial problems, this connection may enable proving correctness of BP variants on a larger class of problems.

From a practical perspective, our theorems proven in section 5 allow us to go beyond the LP relaxation and provably find the MAP even when the LP relaxation is partially fractional. Our experiments on side chain prediction and error correcting code show that using these theorems it is possible to find the MAP on real world instances of very large graphical models where techniques such as junction tree are intractable. Similarly, in our work reported in [14] we have used these theorems to find the global optimum on a number of stereo vision problems. Until our results, only local optima for these problems were known. We believe similar results are possible in a wide range of applications.

### Acknowledgements

Supported by the Israeli Science Foundation. We thank Amir Globerson for comments on a previous version of this manuscript.